\documentclass[letterpaper, 10 pt, conference]{ieeeconf}
\usepackage{cite}
\usepackage{amsmath,amssymb,amsfonts}
\usepackage{graphicx}
\usepackage{hyperref} 
\usepackage{textcomp}
\usepackage{tabularx}
\usepackage{amsmath}
\usepackage{balance}
\usepackage{multirow}
\usepackage{dblfloatfix}
\usepackage{siunitx}
\usepackage{booktabs}
\usepackage{bm}

\usepackage{longtable}
\usepackage{color,colortbl}
\usepackage{algpseudocode}
\usepackage{algorithm}
\definecolor{Gray2}{gray}{0.9}
\definecolor{Gray}{gray}{0.7}
\usepackage{caption}
\usepackage{xcolor}

\usepackage[caption=false,font=footnotesize]{subfig}
\let\labelindent\relax
\usepackage[inline]{enumitem}

\IEEEoverridecommandlockouts % This command is only needed if 
\begin{document}

\title{\LARGE \bf GustPilot: A Hierarchical DRL-INDI Framework for Wind-Resilient Quadrotor Navigation\\
}

\author{Amir Atef Habel, Roohan Ahmed Khan,  Fawad Mehboob, Clement Fortin, and Dzmitry Tsetserukou
\thanks{The authors are with the Intelligent Space Robotics Laboratory, Center for Digital Engineering, Skolkovo Institute of Science and Technology, Moscow, Russia. 
\tt \{Amir.Habel, Roohan.Khan,  Fawad.Mehboob, C.Fortin, D.Tsetserukou\}@skoltech.ru}
% \thanks{Research reported in this publication was financially supported by the RSF-DST grant No. 24-41-02039.}
}
\maketitle

\begin{abstract}
Wind disturbances remain a key barrier to reliable autonomous navigation for lightweight quadrotors, where rapidly varying airflow can destabilize both planning and tracking. This paper introduces \emph{GustPilot}, a hierarchical wind-resilient navigation stack in which a deep reinforcement learning (DRL) policy generates inertial-frame velocity references for gate traversal, while a geometric Incremental Nonlinear Dynamic Inversion (INDI) controller provides low-level tracking with fast residual disturbance rejection. The INDI layer uses incremental feedback on both specific linear acceleration and angular acceleration/rate, relying on onboard sensor measurements to reject wind disturbances during execution. Robustness is achieved through a two-level strategy: wind-aware planning learned via fan-jet domain randomization during training and rapid execution-time disturbance rejection by the INDI tracking controller. We evaluate GustPilot in real flights on a \(50~\mathrm{g}\) quadrotor platform against a DRL--PID baseline across four scenarios ranging from no-wind to fully dynamic conditions with a moving gate and a moving disturbance source. Despite being trained only in a minimal single-gate/single-fan setup, the policy generalizes to more complex environments with up to six gates and four fans without retraining. Across 80 experiments, DRL--INDI achieves an average Overall Success Rate (OSR) of \(\bm{94.6\%}\), compared with \(\bm{36.0\%}\) for DRL--PID, reduces tracking root mean square error (RMSE) by up to \(\bm{50\%}\), and sustains speeds up to \(\bm{1.34~\mathrm{m/s}}\) under wind disturbances up to \(\bm{3.5~\mathrm{m/s}}\). These results demonstrate that combining DRL-based velocity planning with structured INDI disturbance rejection provides a practical approach to wind-resilient autonomous flight.
\end{abstract}


\section{Introduction}
Reliable quadrotor navigation under wind disturbances remains challenging, particularly for lightweight platforms whose low inertia makes them sensitive to moderate gusts. Classical nonlinear and adaptive controllers can provide effective tracking under mild or slowly varying disturbances, but rapidly changing airflow creates coupled planning and control challenges. In such conditions, real-time disturbance modeling is difficult, and tracking performance can degrade.

Recent work shows that learning-based methods can adapt to aerodynamic disturbances beyond classical controllers. For instance, O'Connell et al.~\cite{neuralfly} developed a representation learning framework that adapts quickly to changing wind conditions with formal stability guarantees, holding its position in high-speed wind tunnel tests. Similarly, Huang et al.~\cite{datthuan2023} combined reinforcement learning with adaptive disturbance modeling to track complex trajectories in unsteady wind, outperforming nonlinear controllers and model predictive control.

\begin{figure}[t]
% \begin{figure}[!ht]
\centering
\includegraphics[width=1.0\linewidth]{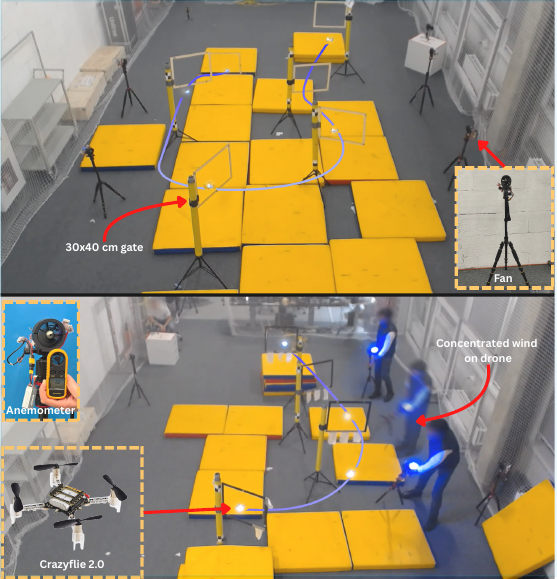} 
\vspace{-0.5cm}
\caption{GustPilot during real-flight gate traversal under distributed and concentrated fan-generated wind disturbances.}
\captionsetup[figure]{skip=0pt} 
\label{fig:title}
\vspace{-0.6cm}
\end{figure}

Despite these advancements, existing methods mainly focus on low-level trajectory tracking or adaptive stabilization, where learning compensates for unknown dynamics. Wind-affected autonomous gate traversal for lightweight quadrotors remains less explored, especially when localized airflow disturbances interact with both planning and tracking. 
This requires wind-aware velocity planning with fast low-level disturbance rejection.

Therefore, we propose GustPilot (Fig.~\ref{fig:title}), a hierarchical DRL--INDI framework that combines a task-specific PPO velocity planner, a geometric INDI low-level controller, and a fan-jet domain-randomized training environment. The policy generates bounded inertial-frame velocity references for gate traversal, while the INDI controller rejects fast wind-induced acceleration and angular-acceleration disturbances during execution. This separation enables a \SI{50}{\gram} Crazyflie-class quadrotor to transfer from minimal single-gate/single-fan training to more complex real-flight scenarios with multiple gates and dynamic fan-generated disturbances.

\section{Related Work}

Recent advances in reinforcement learning have improved autonomous UAV navigation in complex environments. A hybrid approach combining Proximal Policy Optimization (PPO) with Demonstration-Guided Reinforcement Learning (DGRL) and a Control Barrier Function (CBF) action filter demonstrated that formal safety constraints can be incorporated into policy learning while accelerating training via PID-based expert demonstrations \cite{safe_uav_control_demonstration_guided_rl}. Similarly, AgilePilot~\cite{agilepilot} combines DRL with real-time computer vision for high-speed flight in dynamic environments with moving obstacles.In drone racing, learning-based methods have pushed the boundaries of agile flight: Song et al. \cite{DRLracing10.1109/IROS51168.2021.9636053} achieved near-optimal path planning through DRL. Kaufmann et al. \cite{ChampiondroneKaufmann2023} introduced a SWIFT system that uses a PPO-trained policy to generate low-level control commands and a perception system that consists of low-dimensional state observations, hence outperforming professional human pilots. While these works demonstrate substantial agility and safety, they primarily consider nominal conditions without explicitly addressing aerodynamic disturbances such as wind or gusts during policy learning. Consequently, policy robustness against strong external disturbances remains an open challenge.

Several recent studies have applied reinforcement learning to UAV control under aerodynamic disturbances. The impact of sensor noise on sim-to-real transfer was examined in \cite{Sim2real10553074}. Wind-adaptive learning has been explored for dynamic landing and trajectory following, with policies trained in windy conditions showing improved robustness over classical PID control \cite{tornadodrone, landeraiadaptivelandingbehavior, marlander, drl_based_wind_disturbances}. Neural controllers incorporating wind models and adaptation layers have enabled takeoff and landing in strong gusts \cite{neural_controller_takeoff_landing}. Comparisons between model-free and model-based RL in wind \cite{model_free_vs_model_based_icinco24} demonstrated superior tracking performance over traditional controllers, while learning-based low-level controllers proved robust to parameter variations and actuator noise \cite{learningbasedquadcoptercontrollerextreme}. Disturbance-aware frameworks combining RL with stochastic MPC have been proposed to estimate aerodynamic uncertainties while satisfying constraints \cite{constrained_rl_trustworthy}. However, in these approaches, disturbance compensation typically remains decoupled from high-level navigation decisions.

INDI has been employed as an inner loop in different learning-based architectures. In \cite{end_to_end_time_optimal}, end-to-end RL issuing motor commands was compared to a policy generating thrust and body rate commands tracked by an INDI controller, highlighting the robustness benefits of structured inner-loop control. Similarly, \cite{deshmukh2025globalendeffectorposecontrol} integrated PPO-based feedforward with an INDI attitude controller for precise tracking under external forces during aerial manipulation. 

These works show the value of structured INDI inner loops, but they do not address the same combination of velocity-level DRL navigation, geometric INDI tracking, and fan-jet domain randomization for wind-affected gate traversal. GustPilot differs by formulating the learning problem around wind-aware velocity-reference generation while using the INDI layer to reject fast acceleration-level disturbances.

\section{Methodology}

\begin{figure*}[t]
  \centering
  \includegraphics[width=\textwidth]{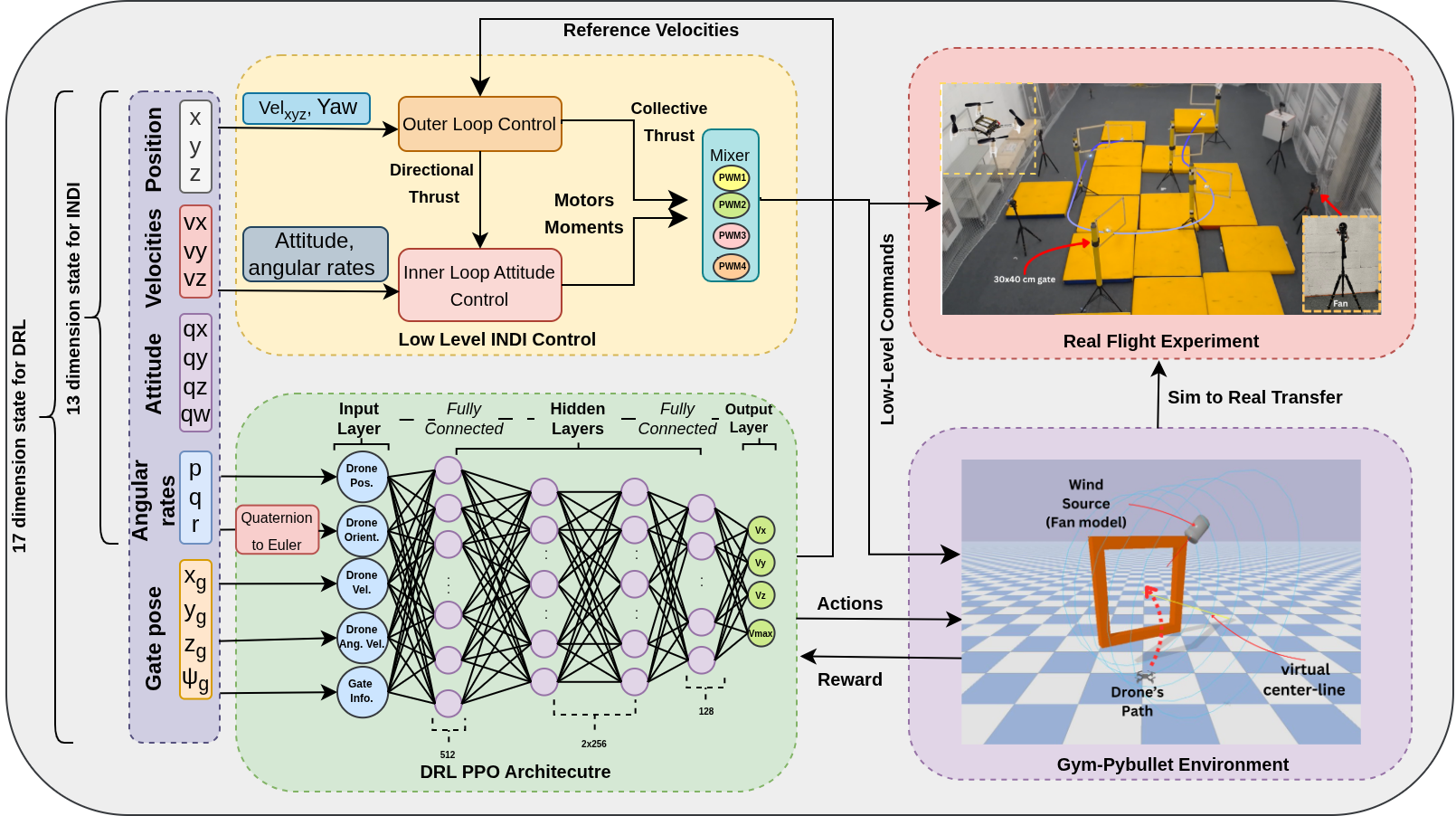}
  % \caption{Overall System Architecture showcasing pipeline of GustPilot.}
  \vspace{-0.5cm}
  \caption{Overview of the GustPilot hierarchical guidance-control architecture.}
  \label{fig:architecture}
  \vspace{-0.6cm}
\end{figure*}

% This section describes the three main components of GustPilot: the geometric INDI controller design used for low-level tracking, the domain-randomized simulation environment, and the PPO policy formulation that maps gate-relative observations to feasible inertial-frame velocity references.

This section describes the three main components of GustPilot: the geometric INDI controller used for low-level tracking, the domain-randomized simulation environment, and the PPO policy formulation that maps gate-relative observations to feasible inertial-frame velocity references.

\subsection{Geometric INDI Control}
\label{subsec:controller_design}

The proposed INDI controller is formulated as a geometric control structure on the Special Euclidean group \(SE(3)\), combining attitude construction on the Special Orthogonal group \(SO(3)\) with measurement-driven incremental inversion. The control architecture consists of (i) an outer loop that regulates the commanded inertial-frame velocity by incrementally updating the specific thrust vector, and (ii) an inner loop that tracks the commanded attitude through angular-acceleration feedback. In this work, the high-level reference consists only of an inertial-frame velocity command \(\mathbf v_{\mathrm{des}}\) and a yaw reference \(\psi_{\mathrm{ref}}\), chosen to align the vehicle heading with the gate normal.

\subsubsection{Sensor Signal Processing}

To reduce sensor noise and bias effects, all raw measurements are filtered using a second-order Butterworth low-pass filters (LPFs), denoted by $\mathcal{F}_{\text{BW}}$. The filtered signals are:
\begin{equation}
\begin{aligned}
\mathbf{a}_f &= \mathcal{F}_{\text{BW}}(\mathbf{a}), \\
\boldsymbol{\omega}_f &= \mathcal{F}_{\text{BW}}(\boldsymbol{\omega}), \\
\dot{\boldsymbol{\omega}}_f &= \mathcal{F}_{\text{BW}}(\dot{\boldsymbol{\omega}}), \\
\boldsymbol{\tau}_{bz,f} &= \mathcal{F}_{\text{BW}}(\boldsymbol{\tau}_{bz}),
\end{aligned}
\end{equation}
where $\mathbf a_f$, $\boldsymbol{\omega}_f$, $\dot{\boldsymbol{\omega}}_f$, and $\boldsymbol{\tau}_{bz,f}$ denote the filtered specific acceleration, angular rate, angular acceleration, and specific thrust vector, respectively; cutoff frequencies are listed in Table~\ref{tab:tuning}. The angular acceleration $\dot{\boldsymbol{\omega}}$ is obtained by numerical differentiation of the raw angular velocity $\boldsymbol{\omega}$ prior to filtering.

\subsubsection{Outer Loop Control}
\label{subsubsec:outer_loop}

The outer loop updates the desired specific thrust vector $\boldsymbol{\tau}_{bz,c}$ incrementally using the measured acceleration:
\begin{equation}
\boldsymbol{\tau}_{bz,c} = \boldsymbol{\tau}_{bz,f} + \alpha_{\text{outer}} \bigl(\mathbf{a}_c - \mathbf{a}_f\bigr),
\end{equation}
where $\alpha_{\text{outer}}\in[0,1]$ is a blending factor, $\mathbf a_f$ is the filtered acceleration measurement, and $\boldsymbol{\tau}_{bz,f}$ is the filtered specific thrust vector from the previous step. In velocity-commanded mode, the commanded acceleration $\mathbf a_c$ is generated purely from velocity tracking:
\begin{equation}
\mathbf{a}_c
=
\mathbf{K}_v\,\mathbf{e}_v
+\mathbf{K}_i \int \mathbf{e}_v\, dt,
\qquad
\mathbf{e}_v = \mathbf{v}_{\text{des}}-\mathbf{v},
\end{equation}
where $\mathbf v_{\text{des}}$ is the inertial-frame velocity reference and $\mathbf v$ is the measured inertial velocity. The diagonal gain matrices $\mathbf K_v$ and $\mathbf K_i$ are given in Table~\ref{tab:tuning}.

The commanded thrust magnitude and direction are:
\begin{equation}
\tau_c = \|\boldsymbol{\tau}_{bz,c}\|,
\qquad
\mathbf{b}_{z,c} = \frac{\boldsymbol{\tau}_{bz,c}}{\tau_c},
\end{equation}
and the collective thrust command is:
\begin{equation}
f_c = m\,\tau_c,
\end{equation}
where $m$ is the quadrotor mass. The unit vector $\mathbf b_{z,c}$ defines the commanded thrust direction used for attitude construction.

\subsubsection{Desired Attitude Computation}

From the commanded thrust direction $\mathbf b_{z,c}$ and yaw reference $\psi_{\text{ref}}$, the desired relative quaternion $\boldsymbol{\xi}_c$ is constructed geometrically. First, the minimum rotation aligning the body $z$-axis $\mathbf e_3$ with $\mathbf b_{z,c}$ is computed. Expressing $\mathbf b_{z,c}$ in the current body frame gives:
\begin{align}
\mathbf{v}_b = \mathbf{R}^T(\mathbf{q}_{\text{cur}})\,\mathbf{b}_{z,c},
\end{align}
and the corresponding ``tilt'' quaternion is:
\begin{equation}
\bar{\boldsymbol{\xi}}_c =
\frac{1}{\sqrt{(1+\mathbf{e}_3\!\cdot\!\mathbf{v}_b)^2+\|\mathbf{e}_3\times\mathbf{v}_b\|^2}}
\begin{bmatrix}
1+\mathbf{e}_3\!\cdot\!\mathbf{v}_b \\
\mathbf{e}_3\times\mathbf{v}_b
\end{bmatrix},
\end{equation}

with $\bar{\boldsymbol{\xi}}_c=[0,1,0,0]^T$ for $\mathbf{v}_b\approx-\mathbf{e}_3$.
% \noindent

\textbf{Yaw reference from the gate normal:}
Let $\mathbf{n}_{\text{gate}}\in\mathbb{R}^3$ denote the unit normal vector of the active gate expressed in the world frame. The yaw reference is defined by aligning the heading with the horizontal projection of $\mathbf n_{\text{gate}}$:
\begin{equation}
\psi_{\text{ref}} = \operatorname{atan2}\!\left(n_{\text{gate},y},\,n_{\text{gate},x}\right).
\end{equation}
Since the gate normal is defined up to sign, the $\pm\;\mathbf{n}_{\text{gate}}$ ambiguity is resolved by flipping it.

The yaw correction quaternion $\boldsymbol{\xi}_\psi$ is computed after applying the tilt correction:
\begin{align}
\mathbf{q}_{\text{int}} &= \mathbf{q}_{\text{cur}} \otimes \bar{\boldsymbol{\xi}}_c, \label{eq:q_int}\\
\mathbf{n}_{\text{ref}} &= [\sin\psi_{\text{ref}},\; -\cos\psi_{\text{ref}},\; 0]^T, \label{eq:n_ref}\\
\bar{\mathbf{n}} &= \mathbf{R}^T(\mathbf{q}_{\text{int}})\,\mathbf{n}_{\text{ref}}, \label{eq:bar_n}\\
\psi &= \operatorname{atan2}\bigl(\bar{n}_x,\,-\bar{n}_y\bigr), \label{eq:psi}\\
\boldsymbol{\xi}_\psi &= \begin{bmatrix} \cos(\psi/2) & 0 & 0 & \sin(\psi/2) \end{bmatrix}^T. \label{eq:xi_psi}
\end{align}
Finally, the full desired relative quaternion combining tilt and yaw is:
\begin{equation}
\boldsymbol{\xi}_c = \bar{\boldsymbol{\xi}}_c \otimes \boldsymbol{\xi}_\psi.
\end{equation}

\subsubsection{Inner Loop Control}
\label{subsubsec:inner_loop}

The inner loop computes torque commands incrementally:
\begin{equation}
\boldsymbol{\mu}_c
=
\boldsymbol{\mu}_f
+\alpha_{\text{inner}}\,\mathbf{J}\bigl(\dot{\boldsymbol{\omega}}_c-\dot{\boldsymbol{\omega}}_f\bigr),
\end{equation}
where $\boldsymbol{\mu}_f$ is the previously applied torque, $\alpha_{\text{inner}}\in[0,1]$ is the blending factor, and $\mathbf J$ is the inertia matrix. The commanded angular acceleration is:

\begin{equation}
\dot{\boldsymbol{\omega}}_c
=
\mathbf{K}_\xi \boldsymbol{\xi}_e
-\mathbf{K}_\omega\boldsymbol{\omega}_f,
\end{equation}
where $\boldsymbol{\xi}_e=\log(\boldsymbol{\xi}_c)$ is the attitude error vector (log map on $SO(3)$), $\mathbf K_\xi$ and $\mathbf K_\omega$ are diagonal gain matrices (Table~\ref{tab:tuning}).

\subsubsection{Control Allocation}
\label{subsubsec:allocation}

The controller outputs the desired wrench $\mathbf w_c=[f_c,\tau_x,\tau_y,\tau_z]^T$, where $f_c$ is the collective thrust (N) and $\boldsymbol{\tau}_c=[\tau_x,\tau_y,\tau_z]^T$ are body torques (N$\!\cdot$m). Motor allocation follows the Crazyflie force--torque mixer:
\begin{equation}
\mathbf w_c = \mathbf A\,\mathbf f,\qquad
\mathbf{A} =
\begin{bmatrix}
1 & 1 & 1 & 1 \\
b & b & -b & -b \\
-b & b & b & -b \\
-c_{\tau} & +c_{\tau} & -c_{\tau} & +c_{\tau}
\end{bmatrix},
\end{equation}
where $\mathbf f=[f_1,f_2,f_3,f_4]^T$ are the per-rotor thrusts, $b=l/\sqrt{2}$ with arm length $l$, and $c_\tau$ is the yaw moment ratio. Saturation enforces actuator limits.

\begin{table}[t]
\centering
\caption{Controller Tuning Parameters and Filter Cutoffs.}
\label{tab:tuning}
\renewcommand{\arraystretch}{1.275}
\setlength{\tabcolsep}{2pt}
\footnotesize
\begin{tabular}{lc}
\toprule
\textbf{Parameter} & \textbf{Value} \\
\midrule
% $K_x$ (outer) & $[5.556,\;5.556,\;12.963]~\mathrm{s}^{-2}$ \\
$K_v$ (outer) & $[3.519,\;3.519,\;31.481]~\mathrm{s}^{-1}$ \\
$K_i$ (outer) & $[0.037,\;0.037,\;5.556]~\mathrm{s}^{-2}$ \\
% $K_a$ (outer) & $[0.037,\;0.037,\;0.037]$  \\
$K_\xi$ (inner) & $[4.643,\;4.643,\;46.08]\times10^9~\mathrm{s}^{-2}$ \\
$K_\omega$ (inner) & $[7.857,\;7.857,\;5.530]\times10^8~\mathrm{s}^{-1}$ \\
$\alpha_{\text{Outer}}$, $\alpha_{\text{Inner}}$ & $1.0$ \\
% $\alpha_{\text{Inner}}$ & $1.0$ \\
Torque saturation & $[0.0,\;0.000949]~\mathrm{N\cdot m}$  \\
Accelerometer LPF & $6$ Hz \\
Gyroscope LPF & $10$ Hz \\
Angular acceleration LPF & $6$ Hz \\
Thrust vector LPF & $10$ Hz \\
% PWM (Min, Max) & ($20000$, $65535$) \\
% PWM to RPM scale $K_{\text{SC}}$ & $0.2685$ \\
% PWM to RPM offset $K_{\text{offset}}$ & $4070.3$ \\
Arm length $(l)$ &   $0.046 ~\mathrm{m}$ \\
Arm‑length factor $(b)$ & $0.03253~\mathrm{m}$ \\
Torque-per-thrust ratio $(c_{\tau})$ &   $0.005964552$ \\
\bottomrule
\end{tabular}
\vspace{-0.5cm}
\end{table}

% -------------------- Simulation Environment --------------------
\subsection{Simulation Environment}
\label{subsec:sim_environment}
% Professor: reject --- adapt
The training environment is implemented as a physics-based simulation. The learning task is intentionally kept simple: single-gate traversal. to minimize confounding factors from complex navigation, allowing us to focus on evaluating disturbance rejection and domain randomization effects in DRL. 
At each episode reset, the simulator randomizes the gate pose, initial drone state, and unobserved wind-field configuration. The parameters \(d_{\min}\) and \(d_{\max}\) define the allowed initial drone-to-gate distance, while the fan-tube geometry and jet parameters in Table~\ref{tab:sim_params} are sampled to cover and slightly exceed the measured real-fan disturbance range. A schematic is shown in Fig.~\ref{fig:architecture}.

\subsubsection{Domain randomization}
The gate pose and drone's initial pose are sampled uniformly from bounded workspaces. The initial drone position is constrained to be placed within a specified distance from the gate (Table~\ref{tab:sim_params}).

\subsubsection{Fan source localization}
The wind-source configuration is defined relative to the gate using a virtual tube aligned with the gate normal. The tube geometry (radius $R_{\mathrm{tube}}$, length $L_{\mathrm{tube}}$) are randomized per episode. Fan sources are sampled around the tube surface and oriented toward its centerline.
% (Table~\ref{tab:sim_params}).

\subsubsection{Jet-Fan Wind Model}
Wind is modeled as a localized turbulent jet. For a point $\mathbf{p}$ relative to a jet origin $\mathbf{o}$ and axis $\mathbf{a}$, the downstream distance $x = (\mathbf{p}-\mathbf{o})^\top\mathbf{a}$, and radial distance $r = \|(\mathbf{p}-\mathbf{o}) - x\mathbf{a}\|$ are computed.

The jet width grows linearly with downstream distance:
\begin{equation}
\sigma(x) = \sigma_0 + k_{\mathrm{spread}} x.
\end{equation}

The centerline speed decays as:
\begin{equation}
u_c(x) = u_0 \frac{x_0}{x_0 + x}.
\end{equation}

The mean axial wind follows a Gaussian radial profile:
\begin{equation}
u_{\mathrm{mean}}(r,x) = u_c(x) \exp\left(-\frac{r^2}{2\sigma(x)^2}\right),
\end{equation}

The wind field is gated to zero outside a cutoff radius $r_{\mathrm{cut}}(x) = \kappa\,\sigma(x)$:
\begin{equation}
\mathbf{v}_{\mathrm{mean}}(\mathbf{p}) = 
\begin{cases}
u_{\mathrm{mean}}(r,x)\,\mathbf{a}, & \text{if } x > 0 \ \wedge\ r \le r_{\mathrm{cut}}(x) \\
\mathbf{0}, & \text{otherwise} .
\end{cases}
\end{equation}

\renewcommand{\labelenumi}{(\alph{enumi})}
\begin{enumerate}

\item{Temporal variability:} Time-correlated turbulence and intermittent gusts are added:
\begin{equation}
\mathbf{v}_{\mathrm{wind}} = \mathbf{v}_{\mathrm{mean}} + \mathbf{v}_{\mathrm{turb}} + \mathbf{v}_{\mathrm{gust}} ,
\end{equation}
where $\mathbf{v}_{\mathrm{turb}}$ is sampled from the 3-D Ornstein--Uhlenbeck (OU) process with correlation time
$\tau_{\mathrm{turb}}$ and scale $\sigma_{\mathrm{turb}}$ \cite{uhlenbeck1930brownian}. The term $\mathbf{v}_{\mathrm{gust}}$
is the intermittent burst activated with low probability and held for a short random duration. 
% (1-cosine shaping can be used)\cite{faa_ac25341}.

\item{Aerodynamic force:} Using a quadratic drag law with relative velocity $\mathbf{v}_{\mathrm{rel}} = \mathbf{v}_{\mathrm{wind}} - \mathbf{v}_{\mathrm{body}}$:
\begin{equation}
\mathbf{F}_w = \frac{1}{2}\rho\,C_dA\,\|\mathbf{v}_{\mathrm{rel}}\|\,\mathbf{v}_{\mathrm{rel}}, \quad \|\mathbf{F}_w\| \le F_{\max}.
\end{equation}

\item{Wind randomization:} Wind is enabled probabilistically per episode. When enabled, the disturbance force is $\mathbf{F}_{\mathrm{jet}}(t;\boldsymbol{\theta}_k)$, where $\boldsymbol{\theta}_k = [u_{0,k},\; F_{\max,k},\; \sigma_{\mathrm{turb},k},\; \tau_{\mathrm{turb},k}]^\top$ is sampled uniformly from predefined ranges (Table~\ref{tab:sim_params}).
\end{enumerate}

\begin{table}[t]
\centering
\caption{Simulation Environment Parameters}
\label{tab:sim_params}
\footnotesize
\renewcommand{\arraystretch}{1.0}
\setlength{\tabcolsep}{2pt}
\begin{tabular}{lll}
\toprule
\textbf{Category} & \textbf{Parameter} & \textbf{Range/Value} \\
\midrule
Domain Rand & $d_{\text{min}}$, $d_{\text{max}}$ (m) & 1.0, 5.0 \\
\midrule
\multirow{2}{*}{Jet Geometry}
    & $R_{\text{tube}}$ (m) & $\mathcal{U}(0.25,1.00)$ \\
    & $L_{\text{tube}}$ (m) & $\mathcal{U}(0.2,1.5)$ \\
\midrule
\multirow{10}{*}{Wind Model} 
    & $u_0$ (m/s) (jet strength) & 1.0–10.0 \\
    & $x_0$ (m) (virtual origin) & 0.20 \\
    & $\sigma_0$ (m) (initial jet width) & 0.10 \\
    & $k_{\text{spread}}$ (spreading rate) & 0.18 \\
    & $\kappa$ (cutoff multiplier) & 3.0  \\
    & $F_{\text{max}}$ (N) & 0.05–1.0 \\
    & $\sigma_{\text{turb}}$ & 0.001–0.20 \\
    & $\tau_{\text{turb}}$ (s) & 0.08–0.40 \\
    & $p_{\text{wind}}$ & 0.5 \\
    & $v_{\text{max}}$ (m/s) (safety clamp) & 12.0 \\
\midrule
Aerodynamic Force & $\rho$ (kg/m³) (air density) & $1.225$  \\
                  & $C_dA$ (m²) (drag-area coeff.) & $0.012$  \\
\bottomrule
\end{tabular}
\vspace{0mm}
\footnotesize
\vspace{-0.5cm}
\end{table}

\subsection{Deep Reinforcement Learning}

\subsubsection{Policy Architecture}

The navigation policy is optimized with PPO \cite{ppocanonical}, but the learning formulation is task-specific: observations encode the drone state and gate-relative geometry, actions are bounded inertial-frame velocity references, and rewards couple gate proximity, frame avoidance, and gate-normal alignment.
The policy takes as input an observation vector that encodes the quadrotor motion state along with task-related information for gate traversal:

\begin{equation}
\mathbf o_t =
\left[
\mathbf x_{\text{drone}},
\mathbf \theta_{\text{drone}},
\mathbf v_{\text{drone}},
\mathbf \omega_{\text{drone}},
\mathbf d_{\text{gate}}
\right],
\end{equation}
where $\mathbf x_{\text{drone}}$ and $\mathbf \theta_{\text{drone}}$ denote the vehicle position and orientation, respectively, while $\mathbf v_{\text{drone}}$ and $\mathbf \omega_{\text{drone}}$ represent the linear and angular velocities. The task-related term $\mathbf d_{\text{gate}}$ contains the relative three-dimensional gate position, size, and orientation.

The actor outputs a velocity-reference action defined as:
\begin{equation}
\mathbf a_t =
\left[
v_x,\,
v_y,\,
v_z,\,
v_{\max}
\right],
\end{equation}
where $v_x$, $v_y$, and $v_z$ represent the desired inertial-frame velocity components, and $v_{\max}$ scales the commanded motion magnitude. The resulting bounded velocity reference $\mathbf v_{\mathrm{des}}$ is sent to the INDI controller, while thrust and torque saturation in the allocation layer further enforce actuator feasibility.

% The policy network consists of fully connected layers with the structure shown in Fig.~\ref{fig:architecture} and stated here as:
The policy network consists of fully connected layers, as shown in Fig.~\ref{fig:architecture}:
\begin{center}
FC512 $\rightarrow$ FC256 $\rightarrow$ FC256 $\rightarrow$ FC128,
\end{center}
with tanh activation functions applied after each hidden layer. The final layers produce the action distribution for the actor and the value estimate for the critic.

\begin{figure}[!t]
\centering
\includegraphics[width=0.9\linewidth]{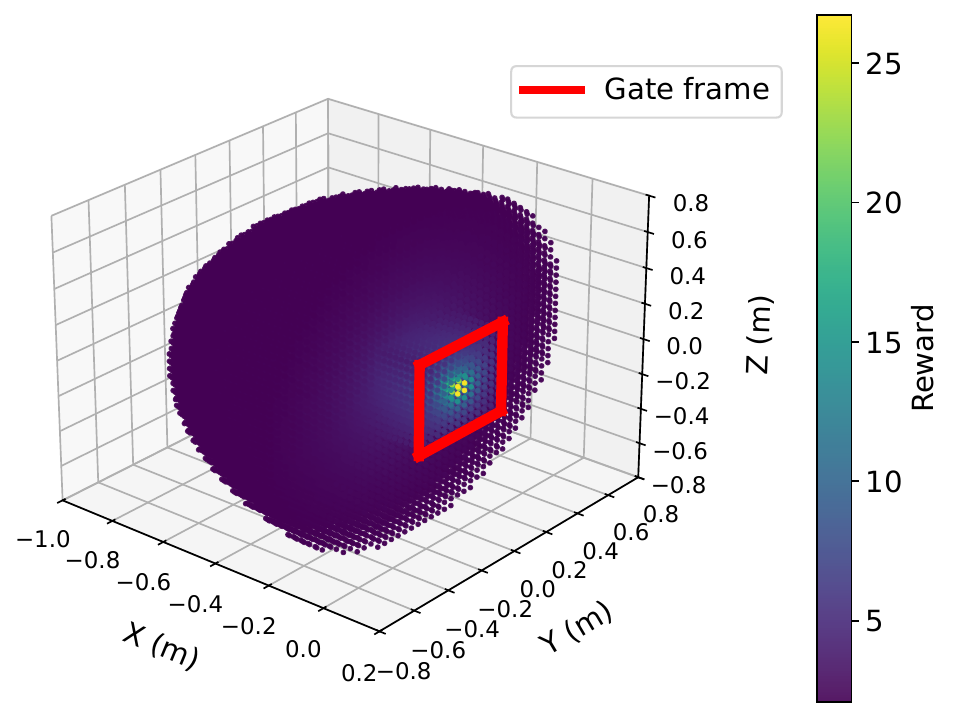} 
\caption{Reward colormap for drone positions relative to the gate.}
\label{reward}
\vspace{-0.55cm}
\end{figure}

\subsubsection{Reward Structure}

The reward is designed to make the policy navigation-focused under wind disturbances, encouraging gate crossing, frame avoidance, and approach alignment rather than direct low-level disturbance compensation.
The main goal is to get as close as possible to the center of the gate, which is achieved by minimizing the distance between the quadrotor and the gate center:

\begin{equation}
R_{\text{proximity}} = \frac{1}{d_{\text{goal}} + c_p}, 
\end{equation}
where $d_{\text{goal}}$ denotes the euclidean distance to the gate and $c_p$ is the small saturation constant that prevents excessively large rewards near the gate center.

A collision penalty is applied when the quadrotor intersects the gate structure:
\begin{equation}
R_{\text{collision}} =
\begin{cases}
- 10 & \text{if a collision occurs}, \\
0 & \text{otherwise}.
\end{cases}
\end{equation}

To encourage smooth traversal, the policy is rewarded for reaching the gate approximately normal to its plane. Let $n_{\text{gate}}$ be the unit normal vector of the gate. The alignment reward is defined as:
\begin{equation}
R_{\text{alignment}} = c_a \, \frac{\mathbf v_{\text{drone}}}{\|\mathbf v_{\text{drone}}\|} \cdot n_{\text{gate}},
\end{equation}
which increases when the velocity direction aligns with the gate normal.

% Finally, the total reward is given by:
The reward coefficients were selected through preliminary randomized simulation trials and then kept fixed for all controllers and scenarios; the total reward is:
\begin{equation}
R_{\text{total}} =
R_{\text{proximity}} +
R_{\text{collision}} +
R_{\text{alignment}}.
\end{equation}
Fig. \ref{reward} visualizes the total reward during a sample step in training in the $X$--$Z$ plane.

\subsubsection{Training}

The policy and value networks are optimized with Adam using the PPO settings in Table~\ref{tab:ppo_params}. Training runs for $4.5\times10^{7}$ steps across $16$ parallel environments, with gate pose, initial drone state, and unobserved wind-field parameters randomized across episodes. The hyperparameters were selected from preliminary simulation trials to favor stable convergence and were kept fixed for all reported controllers and scenarios.

\setlength{\abovetopsep}{0pt}   % space above \toprule
\setlength{\belowbottomsep}{0pt} % space below \bottomrule

\begin{table}[H]
\centering
\setlength{\abovetopsep}{0pt}
\setlength{\belowbottomsep}{0pt}
\caption{Training Parameters}
\label{tab:ppo_params}
\footnotesize
\begin{tabular}{lc}
\toprule
\textbf{Parameter} & \textbf{Value} \\
\midrule
Algorithm & PPO \\
Total training steps & $4.5 \times 10^{7}$ \\
Number of environments & 16 \\
Rollout length & 2048 \\
Batch size & 256 \\
Discount factor ($\gamma$) & 0.99 \\
Clip range & 0.2 \\
Entropy coefficient & 0.001 \\
Learning rate & $1 \times 10^{-5}$ \\
Activation function & Tanh \\
\bottomrule
\end{tabular}
% \vspace{-0.5cm}
\end{table}

\subsection{Training Analysis}

\begin{figure}[t]
\centering
\subfloat[Training reward comparison. The plot shows the evolution of mean episode return over training steps.]{\includegraphics[width=0.48\linewidth]{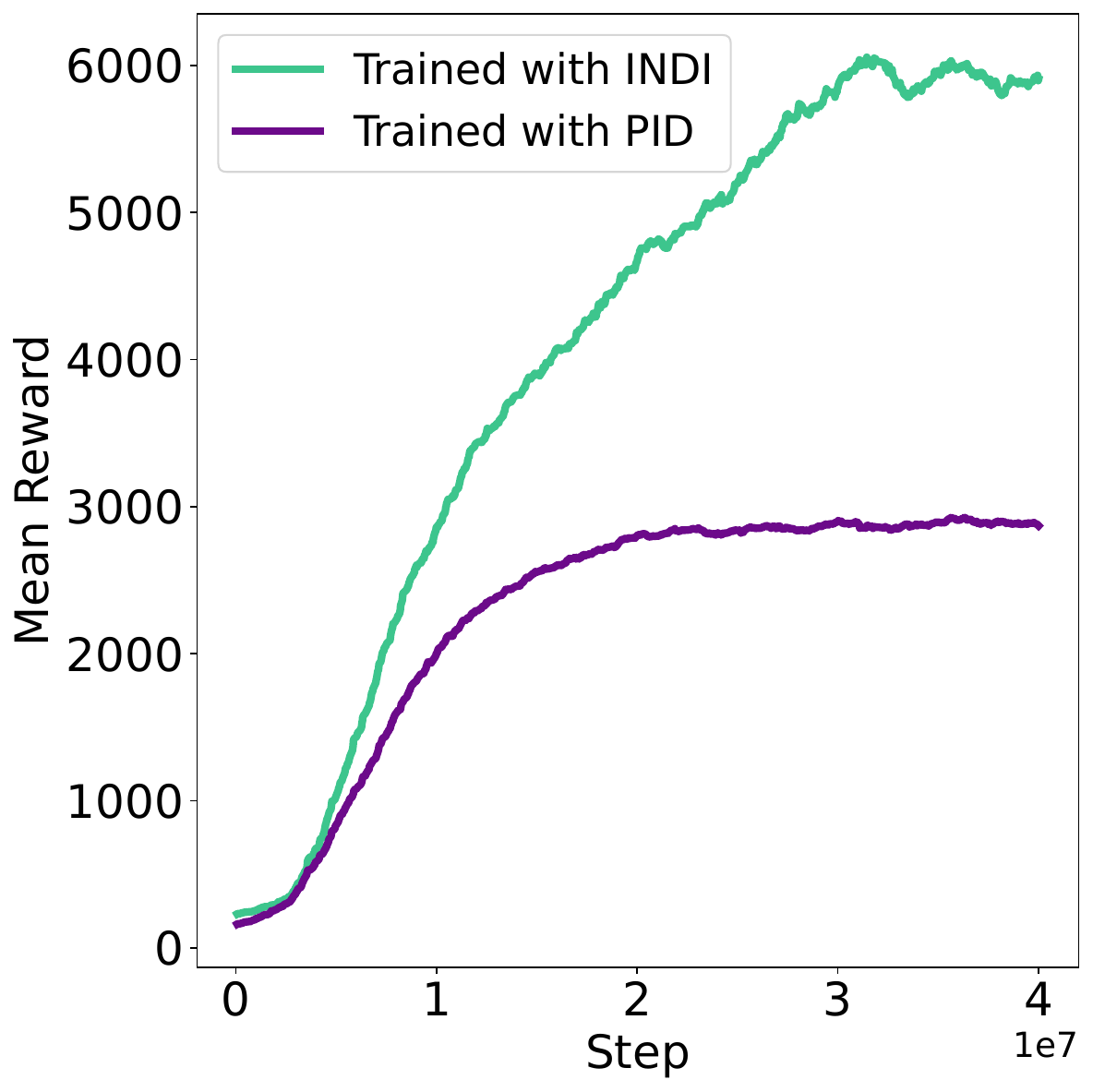}\label{fig:reward_sub}}
\hfill
\subfloat[Comparison of episode duration (s) during training.]{\includegraphics[width=0.48\linewidth]{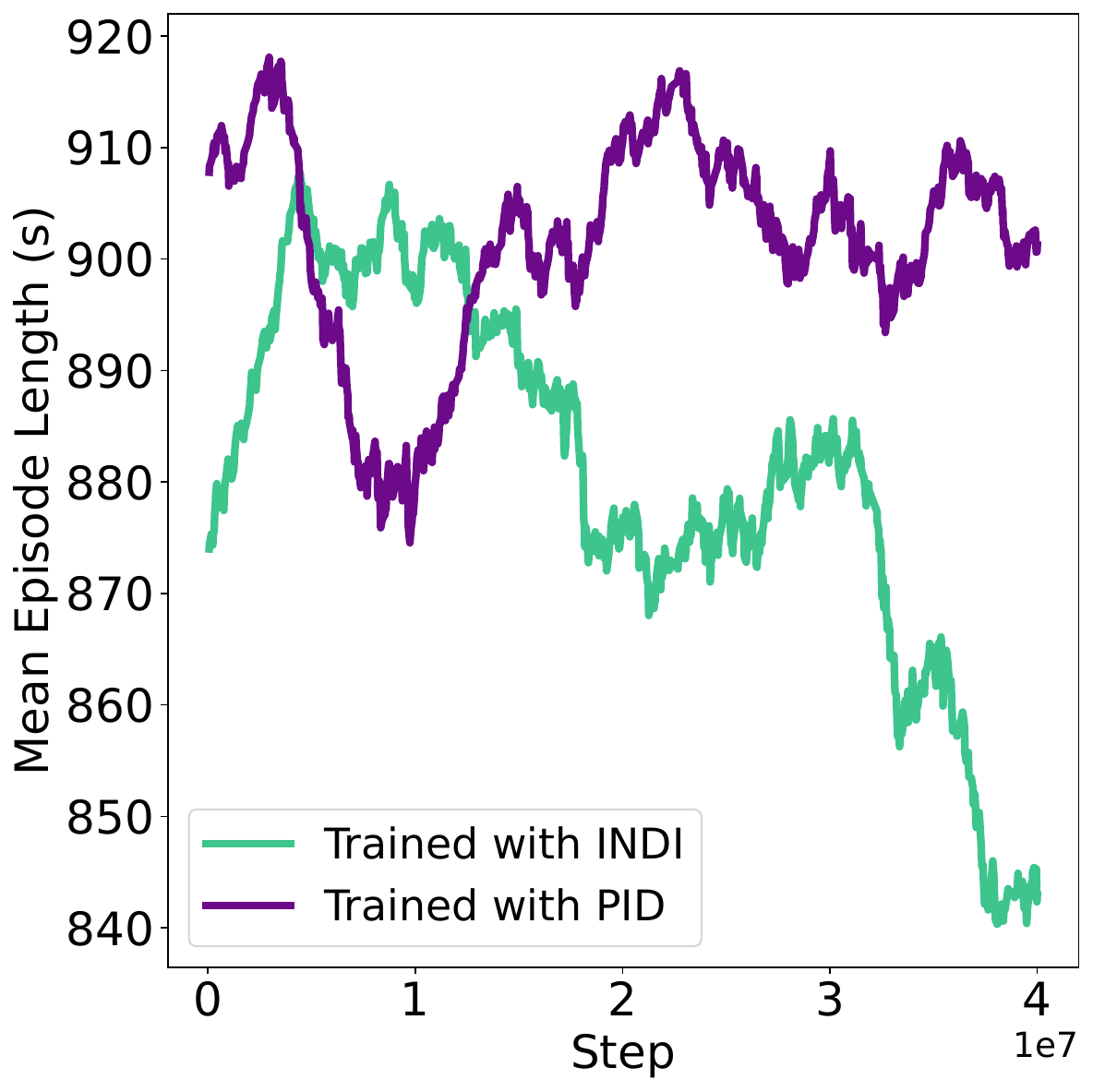}\label{fig:episode_sub}}
\caption{Training metrics for policies learned with INDI and PID low-level controllers.}
\label{fig:training_metrics}
\vspace{-0.205cm}
\end{figure}

The policy trained with the proposed INDI controller consistently achieves higher rewards than the PID-based setup, as shown in Fig.~\ref{fig:reward_sub}. This suggests that improved low-level tracking under wind disturbances allows the learning algorithm to focus more effectively on the navigation objective. 

The geometric INDI controller also yields shorter episode durations during training, indicating faster traversal with fewer early terminations, as shown in Fig.~\ref{fig:episode_sub}.

% \newpage
\section{Flight Experiments}

\subsection{Experimental Setup}

Experiments use a \SI{50}{\gram} Crazyflie 2.1 quadrotor \cite{crazyflies2.0} with our controller running onboard. State estimation is provided by a Vicon motion capture system. The wind disturbances are generated by ducted fans producing measured airspeed up to \SI{7}{m/s} applied to the flight path. The DRL policy runs offboard and sends the inertial-frame velocity reference $\mathbf v_{\mathrm{des}}$ to the onboard controller. The baseline uses the same velocity-command interface with the default Crazyflie PID tracker.
% The DRL policy runs offboard and sends inertial-frame velocity reference $\mathbf v_{\mathrm{des}}$ to the onboard INDI controller, while we used the default PID. 

\subsection{Experimental Scenarios}

Four scenarios illustrated in Fig.~\ref{fig:all_scenarios_2d}(a)--(d) were designed with identical gate dimensions and the same fan model.

\subsubsection{Scenario 1: Nominal Conditions (No Wind)}
Six gates with varying positions and orientations created a trajectory featuring a $1.1\,\mathrm{m}$ altitude change over $0.74\,\mathrm{m}$ and a sharp orientation shift.

\subsubsection{Scenario 2: Distributed Wind Disturbances}
Four stationary fans at arena corners directed airflow toward the flight path, with a minimum distance $0.5\,\mathrm{m}$ to the drone.

\subsubsection{Scenario 3: Concentrated Dynamic Disturbance}
A single fan manually followed the drone's trajectory ($0.43\,\mathrm{m}$ minimum separation, $3.5\,\mathrm{m/s}$ peak airspeed) across four gates to maintain wind exposure along most of the trajectory.

\subsubsection{Scenario 4: Fully Dynamic Environment}
Four gates were placed, with one fan manually guided by a human to follow the drone while the third gate was manually moved horizontally at $0.3\,\mathrm{m/s}$.

% \section{Results}

\subsection{Evaluation Metrics}

We evaluated the proposed DRL--INDI framework against a DRL--PID baseline across four wind scenarios. Both methods use the same observation space, policy architecture, reward, training procedure, and velocity-command interface, so the comparison isolates the low-level tracker: geometric INDI versus PID.

Each controller was tested in ten independent trials per scenario.
Let \(N_{g,s}\) be the number of gates in scenario \(s\), \(N_t\) the number of trials, \(P\) the number of missed gates, \(h\) the number of gate hits, and \(f\) the number of completed trials. The total number of scheduled gate passes is \(G_s=N_{g,s}N_t\). We define:
\begin{align}
S &= 1-\frac{P}{G_s}, &
H &= 1-\frac{h}{G_s}, &
F &= \frac{f}{N_t},
\end{align}
where \(S\) is the Gate Pass Ratio, \(H\) is the Hit-Free Ratio, and \(F\) is the Completion Rate. The Overall Success Rate (OSR) is then:
\begin{equation}
\mathrm{OSR} =
\begin{cases}
\dfrac{F+S+H}{3}, & F>0,\\[0.6em]
0, & F=0.
\end{cases}
\end{equation}
% OSR is used as a compact summary, while the underlying missed gates, hits, and completions are reported separately because \(F\), \(S\), and \(H\) capture different, independent failure modes and thus are weighted equally.
OSR is used only as a compact summary, while the underlying missed gates, hits, and completions are reported separately because \(F\), \(S\), and \(H\) capture different, non-independent failure modes.

% -------- All scenarios (only 2D comparison plots) --------
\begin{figure*}[!t]
\centering
% Row 1: Scenario 1 and Scenario 2
\subfloat[Scenario 1: PID completed the mission with two frame hits at \SI{1.0}{m/s}, whereas INDI achieved a smoother collision-free path at \SI{1.34}{m/s}.]{%
    \includegraphics[width=0.48\textwidth]{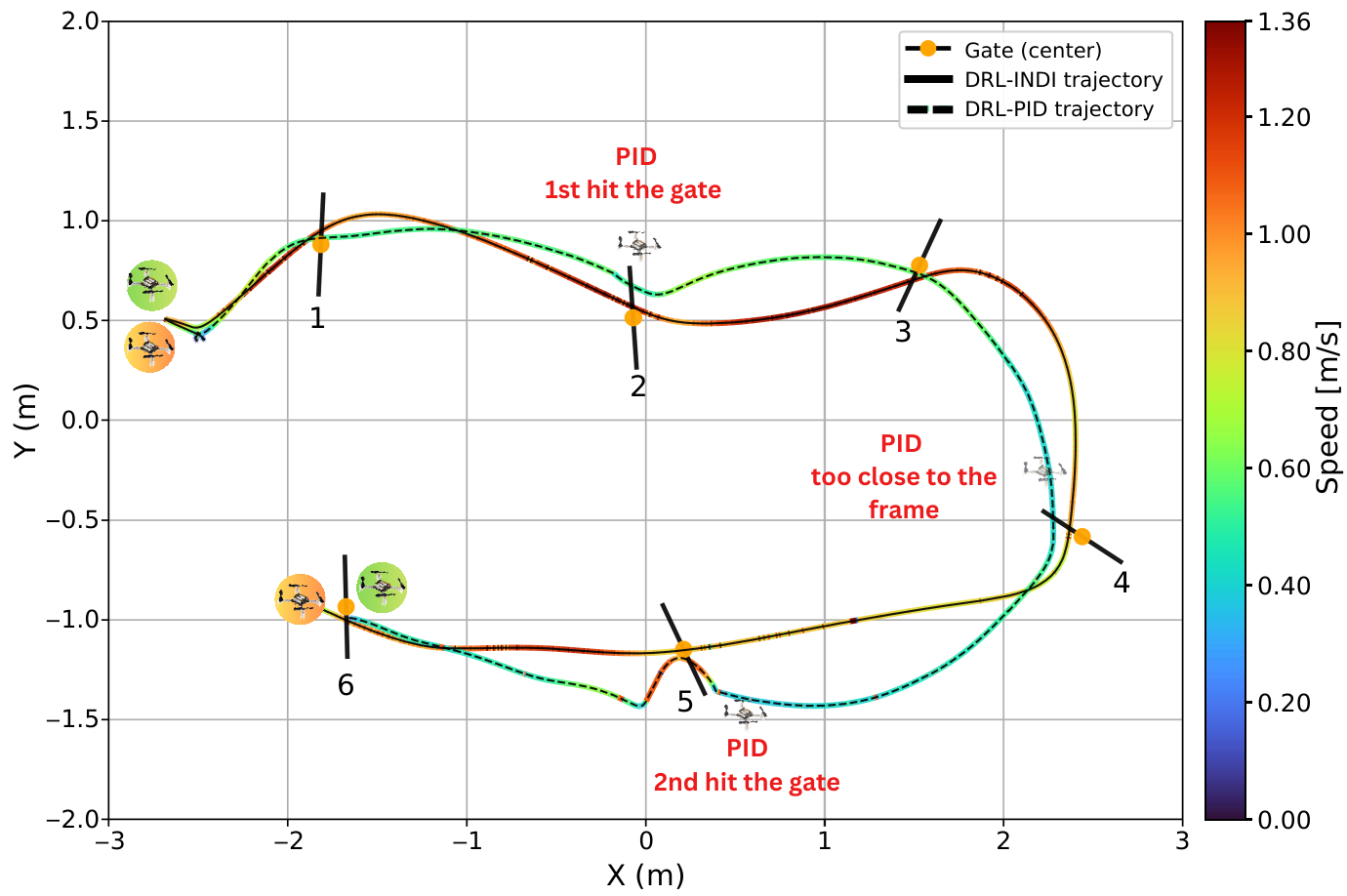}%
}\hfill
\subfloat[Scenario 2: INDI completed the mission at \SI{1.35}{m/s}, whereas PID failed in all trials.]{%
    \includegraphics[width=0.48\textwidth]{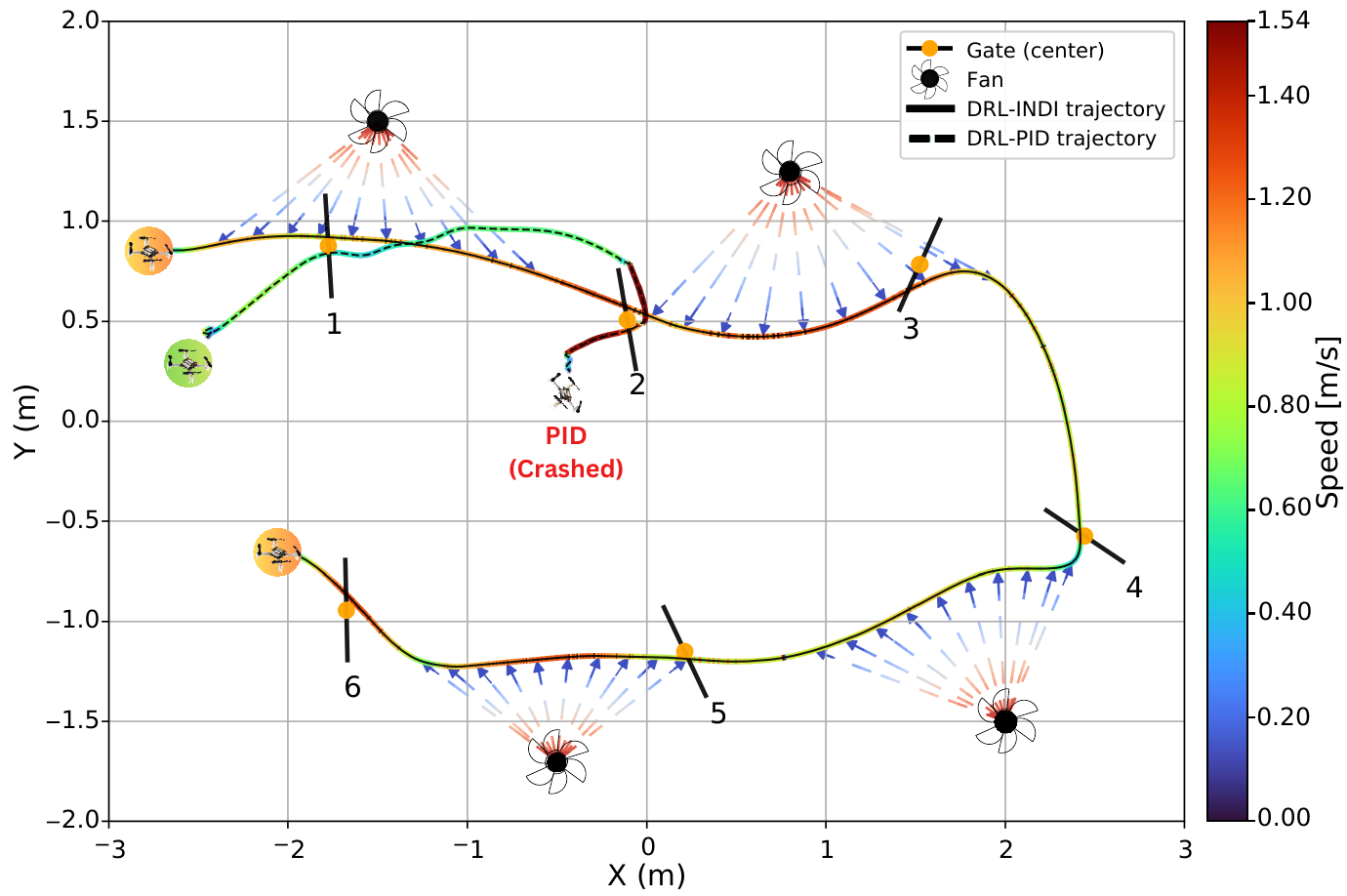}%
}

\vspace{1em}

% Row 2: Scenario 3 and Scenario 4
\subfloat[Scenario 3: PID crashed frequently and reached only \SI{0.9}{m/s}, while INDI completed the mission at \SI{1.12}{m/s}.]{%
    \includegraphics[width=0.48\textwidth]{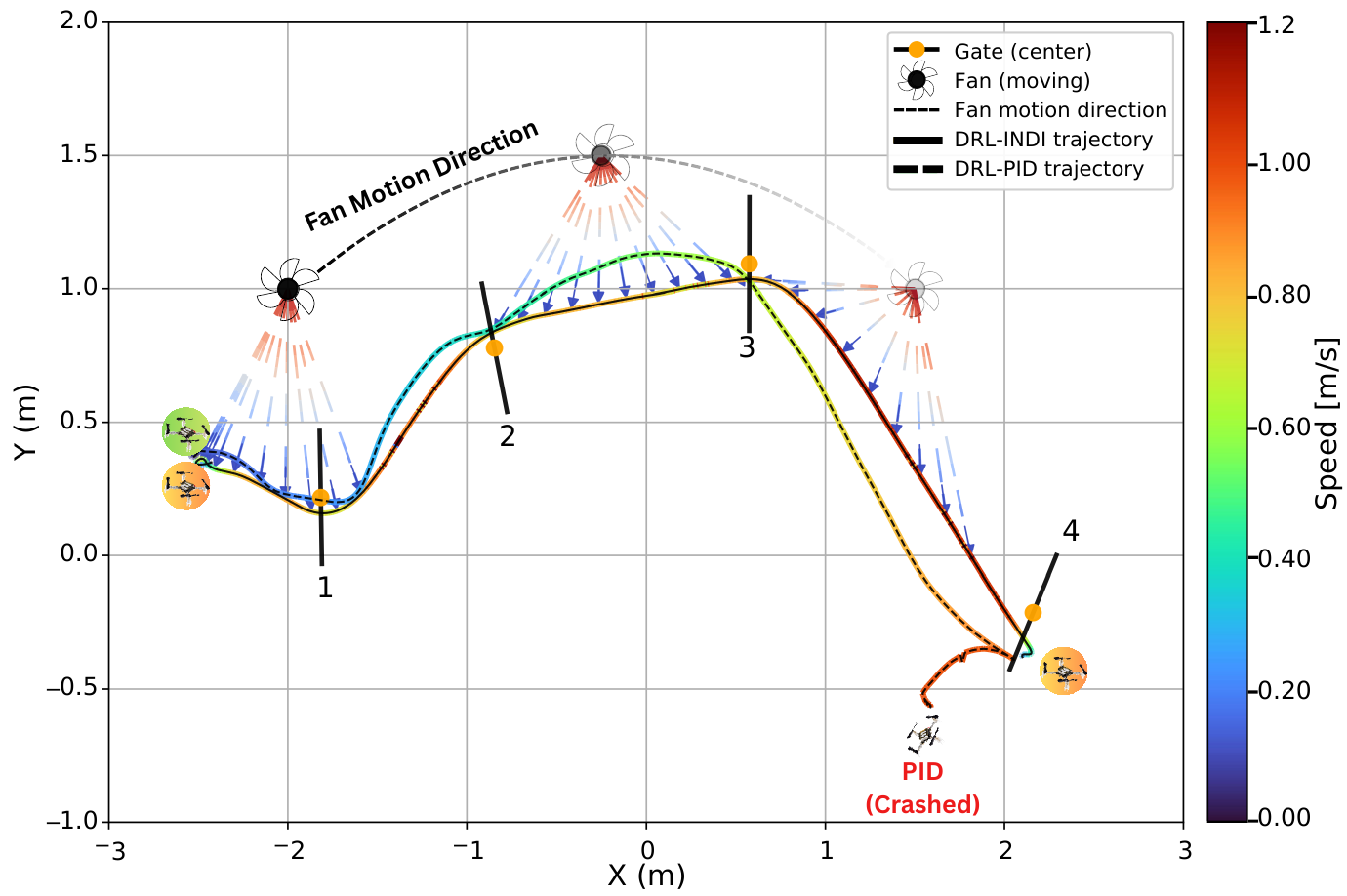}
}\hfill
% \subfloat[Scenario 4: PID failed despite a 35\% speed reduction, whereas INDI completed the mission at \SI{1.3}{m/s}.]{%
%     \includegraphics[width=0.48\textwidth]{figures/scenario44.pdf}
% }
\subfloat[Scenario 4: The PID trajectory shown corresponds to a diagnostic run with a 35\% velocity reduction; nominal-speed PID failed, whereas INDI completed the mission at \(1.3~\mathrm{m/s}\).]{%
    \includegraphics[width=0.48\textwidth]{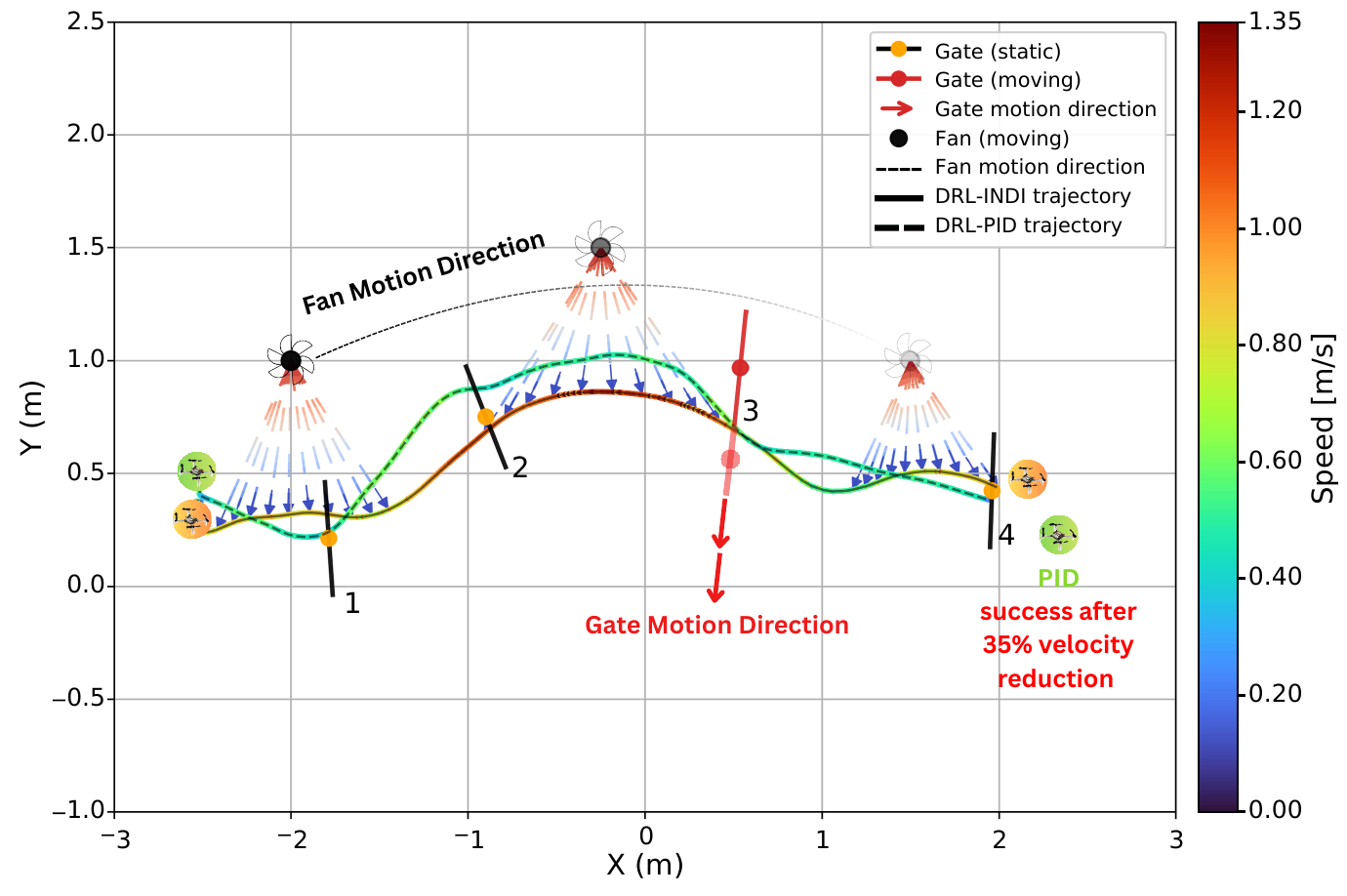}
}

% The PID trajectory shown corresponds to a diagnostic run with a 35\% velocity reduction; nominal-speed PID failed, whereas INDI completed the mission at \(1.3~\mathrm{m/s}\).
\caption{2-D trajectory comparisons for the four real-flight scenarios (DRL--INDI vs. DRL--PID).}
\label{fig:all_scenarios_2d}
\end{figure*}

% -------- Combined results table (corrected labels + consistent with metrics) --------
\begin{table*}[t]
\centering
\caption{Comprehensive Experimental Results: OSR, Missed Gates ($P$), Gate Hits ($h$), Completed Trials ($f$), Closest Approaches, and Tracking Errors.}
\label{tab:combined_results}

\footnotesize
\renewcommand{\arraystretch}{1.1}
\setlength{\tabcolsep}{2pt}

\begin{tabular}{l l c c c c c c c c c c c c c}
\toprule
\textbf{Scenario} & \textbf{Controller} & \textbf{OSR (\%)} & \textbf{$P$} & \textbf{$h$} & \textbf{$f$} & \textbf{Max Speed} & \textbf{Time} & \textbf{Closest Gate} & \textbf{Closest Fan} & \textbf{Airspeed} & \textbf{RMSE} & \textbf{MAE} & \textbf{Max Abs} & \textbf{RMSE Vel.} \\
& & &  &  & & (m/s) & (s) & (m) & (m) & (m/s) & (m) & (m) & (m) & (m/s) \\
\midrule

\multirow{2}{*}{I}
& INDI & 98.8 & 1 & 1 & 10 & 1.337 & 14 & 0.065 & --- & --- & 0.127 & 0.121 & 0.180 & 0.170 \\
& PID  & 88.9 & 6 & 8 & 9 & 1.000 & 24 & 0.096 & --- & --- & 0.190 & 0.220 & 0.210 & 0.200 \\

\midrule

\multirow{2}{*}{II}
& INDI & 97.2 & 2 & 3 & 10 & 1.347 & 14 & 0.040\textsuperscript{a} & 0.700 & 2.44 & 0.119 & 0.111 & 0.184 & 0.163 \\
& PID  & 0.0  & 20 & 20 & 0 & ---   & --- & ---        & ---   & ---  & 1.700\textsuperscript{b} & ---   & ---   & 0.400\textsuperscript{c} \\

\midrule

\multirow{2}{*}{III}
& INDI & 92.5 & 4 & 5 & 10 & 1.122 & 9  & 0.077\textsuperscript{d} & 0.450 & 3.5  & 0.110 & 0.097 & 0.173 & 0.177 \\
& PID  & 58.3 & 12 & 22 & 6 & 0.940 & 14 & 0.130\textsuperscript{d} & 0.976 & 1.67 & 0.145 & 0.131 & 0.211 & 0.182 \\

\midrule

\multirow{2}{*}{IV}
& INDI & 90.0 & 3 & 5 & 9 & 1.311 & 9  & 0.085\textsuperscript{e} & 0.900 & 1.8  & 0.125 & 0.121 & 0.157 & 0.210 \\
% & PID  & 31.7 & 18 & 28 & 1 & 0.680 & 12 & 0.085\textsuperscript{e} & 1.100\textsuperscript{f} & 1.32 & 0.177 & 0.245 & 0.259 & 0.220 \\

& PID  & 0 & 18 & 28 & 0 & 0.680 & 12 & 0.063\textsuperscript{e} & 1.100\textsuperscript{f} & 1.32 & 0.177\textsuperscript{f} & 0.245\textsuperscript{f} & 0.259\textsuperscript{f} & 0.280\textsuperscript{f}\\

\bottomrule
\end{tabular}

\vspace{2mm}
\footnotesize
$G_s=N_{g,s}N_t$ is the total number of scheduled gate passes in scenario $s$ ($N_t=10$ trials). 

Scenarios I--II use $N_{g,s}=6$ ($G_s=60$) and Scenarios III--IV use $N_{g,s}=4$ ($G_s=40$).\\
\textsuperscript{a}Closest gate distance not explicitly reported for Scenario II; RMSE values are provided. \\
\textsuperscript{b}Pre-crash RMSE; the drone failed immediately upon entering the wind field. \\
\textsuperscript{c}Approximate pre-crash velocity RMSE. \\
\textsuperscript{d}Closest pass to gate 1. For PID in Scenario III, the value corresponds to the RMSE at gate 1. \\
\textsuperscript{e}Individual gate distances not reported; RMSE ranges are $0.09$--$0.20\,\mathrm{m}$ for INDI and $0.08$--$0.258\,\mathrm{m}$ for PID (after speed reduction). \\
\textsuperscript{f}Values obtained with 35\% speed reduction; the nominal PID configuration achieved 0\% success in Scenario IV.
\end{table*}

\subsection{Evaluation of Results}

The real-flight performance across the four scenarios is summarized in Table~\ref{tab:combined_results}. In Scenario I, both controllers completed most nominal trials, but DRL--INDI achieved a higher OSR of \(98.8\%\), with fewer missed gates and frame contacts than DRL--PID, as shown in Fig.~\ref{fig:all_scenarios_2d}(a). This shows that the geometric INDI tracker improves trajectory execution even without wind, as also reflected by the tracking-error and speed metrics.

The role of the low-level controller becomes clearer once wind disturbances are introduced. In Scenario II, the distributed fan layout produced sustained disturbances from different directions along the flight path. Since DRL--INDI and DRL--PID use the same PPO policy, observation space, reward, and velocity-command interface, the performance difference can be attributed primarily to the tracking layer. DRL--INDI maintained stable gate traversal with \(97.2\%\) OSR, whereas DRL--PID failed to complete the scenario, as shown in Fig.~\ref{fig:all_scenarios_2d}(b). This indicates that the INDI acceleration-feedback loop preserved tracking authority under sustained wind, while the PID tracker lost disturbance margin.

Scenario III further evaluates the system under a concentrated fan jet intersecting the flight path. DRL--INDI achieved \(92.5\%\) OSR and continued to pass through the disturbed region, where the measured airspeed reached \(3.5~\mathrm{m/s}\), as shown in Fig.~\ref{fig:all_scenarios_2d}(c). In contrast, DRL--PID showed larger deviations and more frequent failures. This result indicates that GustPilot does not rely only on avoiding windy regions. Instead, the PPO policy provides adaptive velocity planning for gate traversal, while the INDI layer compensates for fast local disturbances during execution.

Scenario IV combines two simultaneous challenges: a moving gate and a moving wind source. This case tests whether the policy can continue producing useful velocity references while the low-level controller rejects time-varying aerodynamic disturbances. At the nominal command speed, DRL--PID failed to complete the scenario, whereas DRL--INDI achieved \(90.0\%\) OSR. To diagnose the PID failure mode, an additional PID run was performed with the PPO velocity commands reduced by \(35\%\). This speed-reduced trajectory is shown in Fig.~\ref{fig:all_scenarios_2d}(d) only as a best-effort diagnostic case, not as an equal-speed comparison. The need to reduce PID velocity commands suggests that its failure was mainly due to limited low-level tracking and disturbance-rejection margin under rapidly changing PPO velocity references.

Overall, the results support the central design hypothesis of GustPilot. The PPO policy is responsible for navigation-level velocity planning and adapts the commanded motion according to the observed flight state and gate geometry, while the geometric INDI controller rejects undesirable wind-induced disturbances at the acceleration and angular-acceleration levels. This separation allows a policy trained in a minimal single-gate/single-fan simulation to transfer to more complex real-flight scenarios with multiple gates, distributed wind, concentrated jets, and moving environmental elements.

\section{Conclusion and Future Work}
This paper presented GustPilot, a hierarchical DRL--INDI framework for wind-resilient quadrotor gate traversal. GustPilot combines a PPO-based velocity planner with a geometric INDI low-level controller, separating navigation-level decision making from fast disturbance rejection. Across four real-flight scenarios, DRL--INDI consistently outperformed the DRL--PID baseline, achieving higher success, fewer missed gates and frame contacts, and lower tracking errors. The framework maintained an OSR of \(\geq 90.0\%\) across all scenarios, with drone speeds up to \(1.347~\mathrm{m/s}\) and fan-generated wind disturbances up to \(3.5~\mathrm{m/s}\). The largest gaps appeared under sustained and concentrated disturbances, where PID lost disturbance margin while INDI maintained stable tracking through acceleration and angular-acceleration feedback.

The results support the central design principle of GustPilot: the PPO policy performs navigation-level velocity planning from the observed vehicle state and gate geometry, while the low-level INDI controller rejects fast wind-induced disturbances during execution. This separation reduces the burden on the learned policy and enables transfer from a minimal single-gate/single-fan training setup to more complex real-flight scenarios with multiple gates, distributed wind, concentrated jets, and moving environmental elements. Thus, the policy adapts the commanded motion, while the structured controller preserves tracking authority against low-level aerodynamic effects.

These findings also define directions for further validation. The learning formulation is mission-specific to wind-affected gate traversal, with observations, actions, and rewards defined around gate-relative navigation rather than general autonomous exploration. The real-flight experiments are conducted in an indoor motion-capture arena, where Vicon provides state feedback and the DRL policy runs offboard. The controller and motor allocation are implemented for a \(50~\mathrm{g}\) Crazyflie platform, so transfer to larger UAVs requires retuning and validation under different inertia, actuator bandwidth, thrust-margin, and disturbance-to-weight conditions.

Building on these results and studying the current limitations,  the future work will extend GustPilot toward online wind-adaptive navigation. We will investigate residual wind-force estimation to identify persistent disturbances and adapt the velocity planner, training distribution, or INDI gains. We will also study tighter policy-controller interfaces, such as residual force or torque corrections, while preserving the stability and disturbance-rejection advantages of the INDI inner loop. Finally, replacing Vicon-based feedback with onboard state estimation and evaluating larger UAV platforms will be necessary to assess deployment beyond indoor motion-capture settings.

\section*{Acknowledgments} 
Research reported in this publication was financially supported by the RSF grant No. 24-41-02039.

% \newpage
\bibliographystyle{IEEEtran}
\bibliography{bibliography}
\balance
\addtolength{\textheight}{-12cm}
\end{document}